\documentclass[11pt]{article}

\usepackage[preprint]{acl}

\usepackage{times}
\usepackage{latexsym}

\usepackage[T1]{fontenc}

\usepackage[utf8]{inputenc}

\usepackage{microtype}

\usepackage{inconsolata}

\usepackage{graphicx}
\usepackage{amsfonts}
\usepackage{subcaption}
\usepackage{enumitem}

\usepackage[most]{tcolorbox} 

\usepackage{booktabs}   
\usepackage{multirow}   
\usepackage{array}      
\usepackage[table]{xcolor} 
 
\newcommand{\negat}[1]{\textcolor[HTML]{FF0000}{\scriptsize{(-#1)}}}

\newtcbtheorem[]{exmp}{Prompt}%
{colback=gray!5,colframe=gray!75!black,fonttitle=\bfseries, left=.02in, right=.02in,bottom=.02in, top=.02in}{exmp}

\usepackage{fontawesome5} 

\definecolor{hy_planner}{RGB}{0, 85, 128}    
\definecolor{hy_exec}{RGB}{34, 139, 34}      
\definecolor{hy_branch}{RGB}{106, 27, 154}   
\definecolor{hy_mem}{RGB}{216, 67, 21}       
\definecolor{hy_bg}{RGB}{248, 249, 250}      

\newcommand{\think}[1]{{\color{hy_planner}\small\textbf{\texttt{<think>}}} \textit{#1} {\color{hy_planner}\small\textbf{\texttt{</think>}}}}
\newcommand{\task}[1]{{\color{hy_planner}\small\textbf{\texttt{<task>}}} \textsf{#1} {\color{hy_planner}\small\textbf{\texttt{</task>}}}}
\newcommand{\result}[1]{\vspace{2pt}\noindent\hspace{1em}{\color{hy_exec}\small\textbf{\texttt{<result>}}} #1 {\color{hy_exec}\small\textbf{\texttt{</result>}}}}
\newcommand{\branch}[1]{\vspace{2pt}\noindent\hspace{1em}{\color{hy_branch}\small\textbf{\texttt{<isoReason>}}} #1 {\color{hy_branch}\small\textbf{\texttt{</isoReason>}}}}

\newcommand{\memcompress}[1]{\vspace{4pt}\noindent\fcolorbox{hy_mem}{white}{\parbox{0.96\linewidth}{\centering \color{hy_mem}\small \faCompress\ \textbf{Memory Compression}: #1}}\vspace{4pt}}

\title{HyMem: Hierarchical Context Management for Long-Horizon Agents via Information Isolation}

\author{
XinQi Wang\textsuperscript{1,2,3,*},
Jinwei Xiao\textsuperscript{1,2,*},
Sijia Cui\textsuperscript{1,2},
Hongming Zhang\textsuperscript{1},
Yanna Wang\textsuperscript{1},
Qingyang Zhang\textsuperscript{1},
Bo Xu\textsuperscript{1,3}
\\
\textsuperscript{1}National Key Laboratory of Cognition and Decision Intelligence for Complex Systems, \\ Institute of Automation, Chinese Academy of Sciences
\\
\textsuperscript{2}University of Chinese Academy of Sciences
\\
\textsuperscript{3}Nanjing Artificial Intelligence Research of IA
\\
\small{
\textsuperscript{*}Equal Contribution
}
\\
\small{
\texttt{wangxinqi2024@ia.ac.cn}
}
}

\begin{document}
\maketitle

\begin{abstract}
Large language model (LLM) agents often perform poorly on complex, long-horizon tasks because their context becomes increasingly cluttered over time. As interactions accumulate, detailed execution traces and intermediate outputs dominate the context, making it difficult for the model to retain and use high-level planning information. Most existing methods address this issue through compression or retrieval applied to a single, flat context, which does not clearly separate different types of context information and often leads to degraded reasoning. To address this challenge, we propose HyMem, a hierarchical framework that explicitly separates the agent’s context into distinct functional layers. HyMem organizes context by function to separate high-level planning from execution and complex analysis. Its isolated reasoning module handles complex subtasks without adding intermediate reasoning traces to the persistent planning context, while its memory management module preserves task progress across context refreshes through structured summaries. These components reduce redundant context accumulation, retain task-critical information, and support coherent long-horizon reasoning within a limited context window.
Experiments on GAIA and Browsecomp-plus show that, with DeepSeek-V4, HyMem achieves average Pass@1 scores of 66.7\% and 61.3\%, outperforming the strongest baseline by 6.1 and 4.7 percentage points, respectively. Further analysis indicates that HyMem effectively controls the growth of the reasoning context, allowing the model to maintain focus and accuracy across complex, long-horizon tasks.
\end{abstract}

\section{Introduction}
\label{sec:introduction}

\begin{figure}[t]
    \includegraphics[width=\columnwidth]{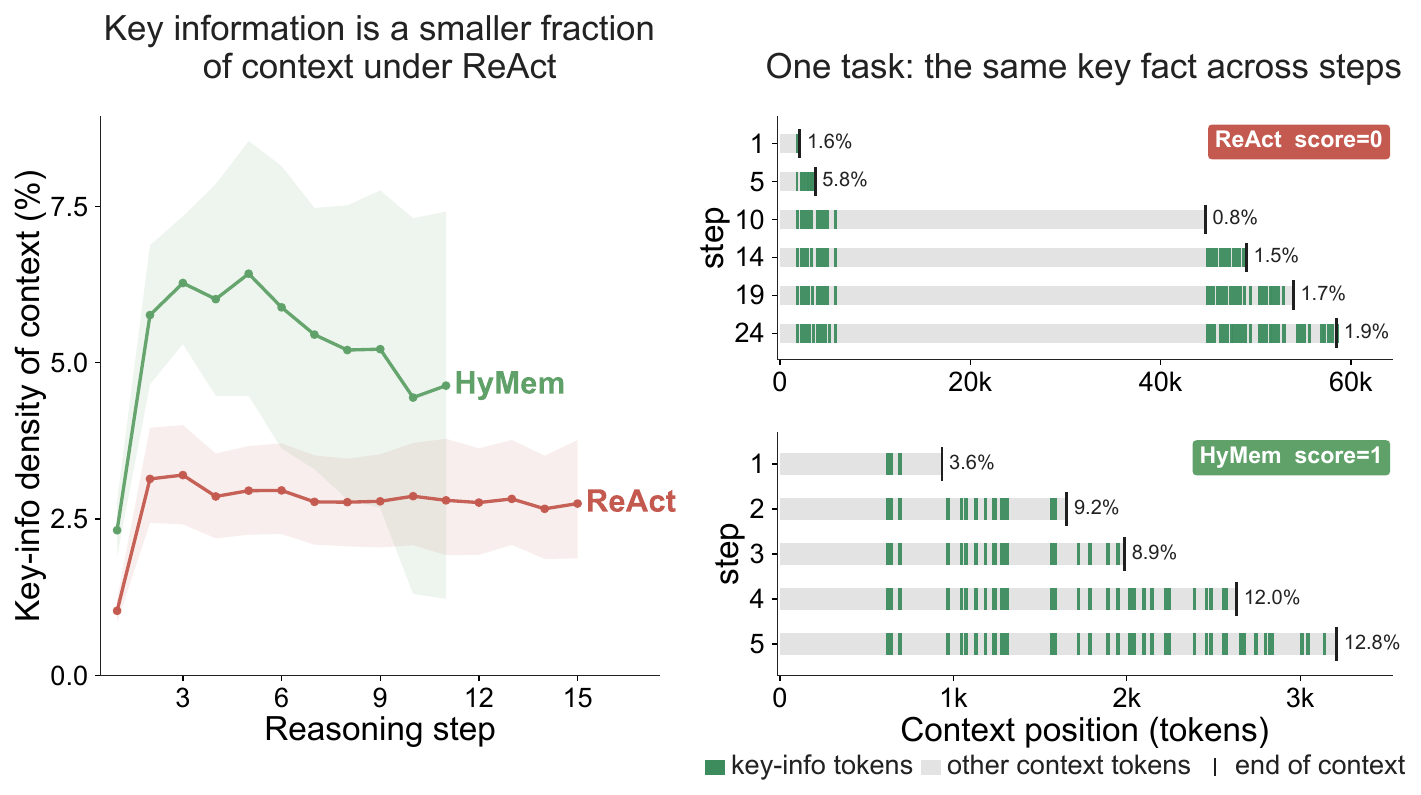}
    \caption{\textbf{Changes in task-critical information within the agent context.} As the context length increases, effective task signals are diluted by redundant historical information, reducing the agent’s sensitivity to core constraint information. The left panel shows changes in task-critical information during task execution on GAIA by agents using ReAct and HyMem, respectively. The right panel illustrates the changes in the position and density of ReAct and HyMem within the context in a task example.}
    \label{fig:task_info}
\end{figure}

Large Language Model-based agents have demonstrated remarkable capabilities in tackling intricate, long-horizon tasks across scenarios requiring extensive environmental interaction, such as graphical user interface manipulation~\citep{qin2025ui,tang2025agentkbleveragingcrossdomain}, in-depth research investigation~\citep{wei2025browsecompsimplechallengingbenchmark,jin2025searchr1trainingllmsreason,DBLP:journals/corr/abs-2501-05366,openaideepresearchcard,google2025deepresearch}, and web-based information retrieval~\citep{zhou2023webarena,Li2025WebThinker,qiao2025webresearcherunleashingunboundedreasoning,cui-etal-2025-self}. However, when facing scenarios that mirror the complexity of real-world tasks---necessitating a seamless integration of macro-strategic planning and micro-tool manipulation---LLM-based agents still encounter significant challenges. A critical bottleneck emerges from the structural mismatch between the agent's cognitive processes and its context management. Long-horizon and complex tasks inherently involve different information streams: high-level strategic planning which requires stability and continuity, low-level execution traces which are often noisy and transient~\citep{zhao2024longagentscalinglanguagemodels,yu2025memagentreshapinglongcontextllm,yang2025learningjobexperiencedrivenselfevolving,zhao2025llmbasedagenticreasoningframeworks,zhang2024a}. In standard flat context architectures (e.g., ReAct~\citep{yao2023react}), these distinct streams are interleaved into a single sequence. As shown in Figure~\ref{fig:task_info}, the redundancy information from execution steps rapidly accumulates, creating an "information flood" that dilutes and eventually submerges the sparse but critical signals of the strategic plan. This leads the agent to lose track of its original goal amidst the details of tool usage.

Prior approaches have attempted to mitigate context constraints through techniques such as context compression~\citep{yan2025memoryr1enhancinglargelanguage,zhang2025memgenweavinggenerativelatent,Yang_2024,wang2025memalphalearningmemoryconstruction,2025mem1learningsynergizememory} or external memory retrieval~\citep{yang2026coarsetofinegroundedmemoryllm,leng2024longcontextragperformance,xu2025mem,li2025deepagentgeneralreasoningagent,chhikara2025mem0buildingproductionreadyai,zhang2023replay}. However, these methods typically operate on a pre-mixed flat sequence, making it difficult to distinguish the underlying reasoning signal from the execution noise. Specifically, compression often leads to the loss of critical reasoning chains, and external memory retrieval frequently fails to preserve temporal logic and struggles to differentiate between high-level strategic planning and discrete execution steps. Ultimately, these strategies merely address the symptom of context length rather than the root cause: the way different types of information are mixed together in a single sequence.

\begin{figure*}[ht] 
    \centering
    \includegraphics[width=0.95\textwidth]{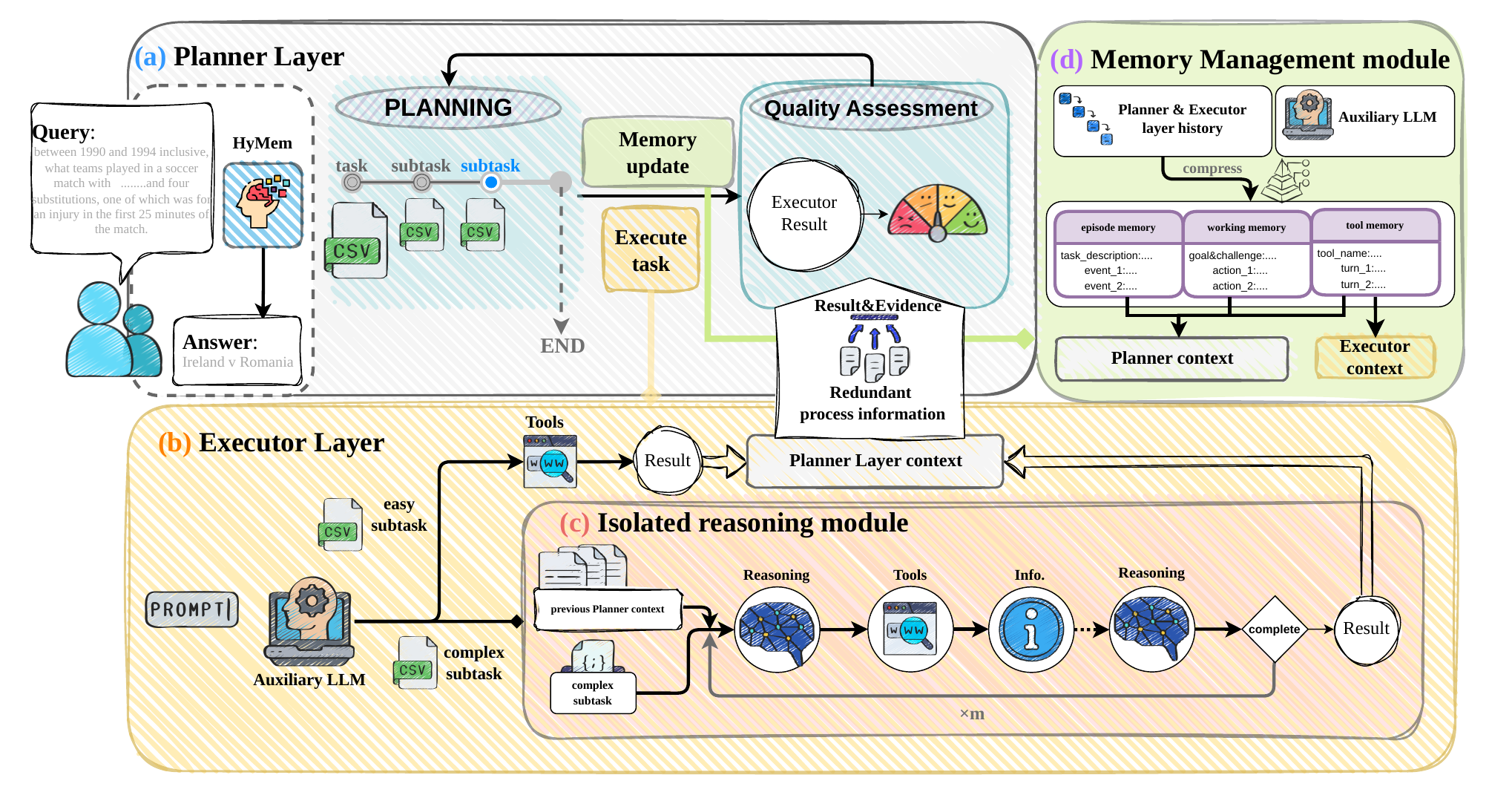} 
    \caption{Overview of the HyMem framework. Upon query arrival, the PLanner Layer (a) performs task decomposition and iterative planning. Sub-tasks are routed to the Executor Layer (b) for execution within isolated reasoning spaces (c) based on complexity, returning refined results for strategic synthesis. Simultaneously, the framework maintains context efficiency by compressing historical trajectories into structured episode, working, and tool memories in (d), which are re-injected to ensure long-horizon task continuity and reasoning focus.}
    \label{fig:hymem_architecture}
\end{figure*}

To address the challenges, we propose HyMem, a framework that enforces hierarchical management and isolation between reasoning layer and execution layer. Instead of treating context as a passive log of past interactions, HyMem manages it deliberately to keep only what is useful for decision-making. By decoupling high-level strategic planning from low-level execution details, the framework encapsulates noisy interaction loops within a dedicated Action Layer, ensuring that the Planner Layer context remains clean and focused on high-value strategic signals. This source-level isolation fundamentally blocks the propagation of redundant execution information into the decision-making space, allowing for sustained reasoning clarity. Complementing this, HyMem uses a structured compression method to summarize past interactions into concise representations, reducing context size while preserving essential task information. As a training-free solution for the long-horizon reasoning task, HyMem offers seamless, plug-and-play convenience for complex, long-horizon tasks.

To validate the effectiveness of the HyMem framework, we conducted systematic evaluations on two representative benchmark datasets, covering complex information retrieval tasks as well as general-purpose agent reasoning tasks. HyMem improves average Pass@1 over the strongest baseline by 6.1 percentage points on GAIA and 4.7 percentage points on Browsecomp-plus~\citep{wu2025resumunlockinglonghorizonsearch,lu2025scalingllmmultiturnrl,2025mem1learningsynergizememory,yu2025memagentreshapinglongcontextllm}. On hard Browsecomp-plus tasks, Pass@1 increases from 14.0\% to 30.0\%. Ablation studies show that isolated reasoning is important for long-horizon tasks: removing this module consistently lowers task success, indicating that complex subtask analysis should remain separate from the persistent planning context. Structured memory serves a complementary role by preserving task goals, completed milestones, and reusable tool experience across context refreshes. These findings suggest that effective long-horizon context management requires both selective information isolation and continuity preservation, rather than context compression alone. Furthermore, analysis of context length trajectories reveals that the HyMem architecture's global specification inference capability is less affected by redundant information during execution, demonstrating its potential for sustained reasoning and extensive exploration in complex, long-threaded tasks.

In summary, our contributions are as follows:
\begin{enumerate}
    \item[(1)]We formalize context dilution in long-horizon agents as an information-density mismatch between sparse planning signals and dense execution traces, and introduce hierarchical context isolation as the guiding principle.
    \item[(2)]We instantiate this principle in HyMem, a training-free framework that separates planner, executor, isolated subtask reasoning, and structured memory management through explicit information boundaries.
    \item[(3)]We evaluate HyMem on Browsecomp-plus and GAIA against  several methods. HyMem achieves average Pass@1 scores of 61.3\% and 66.7\%, outperforming the strongest baseline by 4.7 and 6.1 percentage points, respectively, while maintaining comparable token consumption and tool-call counts.
\end{enumerate}

\section{Related Work}
\label{sec:related_work}

\noindent \textbf{LLM-based Agent Frameworks.}
LLM-based agents have been widely studied as systems that combine language-model reasoning with tool use and environmental interaction~\citep{wang2023survey,comanici2025gemini25pushingfrontier,jin2022buildgenerallyreusableagentenvironment}. Representative frameworks such as ReAct~\citep{yao2023react}, Plan-and-Solve~\citep{wang-etal-2023-plan}, and Reflexion~\citep{NEURIPS2023_1b44b878} improve task solving by making intermediate reasoning, action selection, planning, or self-reflection explicit. Recent agent systems further extend this paradigm to web search, GUI control, and long-horizon information-seeking tasks~\citep{yang2025learningjobexperiencedrivenselfevolving,wu2025webdancerautonomousinformationseeking,li2025websailorv2bridgingchasmproprietary}. Closely related to our work, THREAD~\citep{schroeder-etal-2025-thread} explores recursive child threads for deeper subtask reasoning. These methods demonstrate the value of structured reasoning workflows, but they mainly focus on how agents decompose, execute, or reflect on tasks. HyMem instead focuses on how information should be separated during execution: high-level planner, tool interaction, and deep subtask reasoning are placed in different context spaces, and only synthesized results are returned to the main decision context.

\noindent \textbf{Agent Memory and Context Management.}
Another line of work improves long-horizon agents through memory construction, retrieval, summarization, or learned context management~\citep{mei2025surveycontextengineeringlarge}. External memory systems such as A-MEM~\citep{xu2025mem}, Mem0~\citep{chhikara2025mem0buildingproductionreadyai}, and MemOS~\citep{li2025memosoperatingmemoryaugmentedgeneration} organize reusable information outside the immediate context and retrieve relevant memories when needed. Intra-task context management methods such as ReSum~\citep{wu2025resumunlockinglonghorizonsearch}, MemAgent~\citep{yu2025memagentreshapinglongcontextllm}, MEM1~\citep{2025mem1learningsynergizememory}, AgentFold~\citep{ye2025agentfoldlonghorizonwebagents}, and context-folding approaches~\citep{sun2025scalinglonghorizonllmagent} compress or learn to retain useful trajectory information during long tasks. These approaches reduce context pressure, but they usually operate on trajectories where planning signals and execution traces have already been mixed, or require additional training to learn memory operations. HyMem is complementary to these methods: it uses structured memory to preserve task continuity, but its central mechanism is architectural context isolation, which reduces the amount of low-level execution noise that reaches the planner context in the first place.

\section{Methodology}
\label{sec:methodology}

As illustrated in Figure~\ref{fig:hymem_architecture}, HyMem is a training-free inference-time controller for long-horizon LLM agents. Its central design principle is typed context isolation: planning, tool execution, isolated sub-task reasoning, and memory consolidation are maintained in separate context spaces, and only schema-constrained messages are allowed to cross their boundaries. This design prevents raw execution traces and intermediate deliberation tokens from being directly appended to the main planning context, while preserving task continuity through structured memory.

\subsection{Problem Formulation}
\label{sec:hymem-problem}

We consider an LLM-based agent that solves a user query $q$ through iterative interaction with an environment $\mathcal{E}$. At step $t$, the agent emits an action or reasoning output $a_t$ and receives an observation $o_t \in \mathcal{O}$. A conventional ReAct-style agent maintains a flat context:
\begin{equation}
    C_t = \big(P, q, (a_1,o_1), \dots, (a_t,o_t)\big),
\end{equation}
where $P$ is the system prompt and all items are concatenated as tokens.

The information in $C_t$ is heterogeneous. We use two functional projections to describe this structure:
\begin{equation}
    \begin{aligned}
    Z_t = \pi_Z(C_t), \\  X_t = \pi_X(C_t).
    \end{aligned} 
\end{equation}

where $\pi_Z(\cdot)$ denotes a conceptual selector that extracts planning-relevant information from the context, $\pi_X(\cdot)$ denotes a selector for execution-level traces. $Z_t$ contains planning-relevant information such as goals, verified facts, constraints, and unresolved sub-goals, while $X_t$ contains execution-level traces such as raw tool outputs, browsing logs, retries, formatting artifacts, and failed attempts. In long-horizon tasks, $|X_t|$ typically grows faster than $|Z_t|$, reducing the relative density of planning-relevant information:
\begin{equation}
    \rho_t = \frac{|Z_t|}{|C_t|}.
\end{equation}
We refer to this reduction ascontext dilution. Figure~\ref{fig:task_info} empirically measures this phenomenon by tracking task-critical information across reasoning steps.

HyMem aims to update the planner only through distilled typed messages. Let $R_t$ denote the structured return from a lower-level context space. The intended information boundary is:
\begin{equation}
\label{eq:info-constraint}
\begin{aligned}
    C_{t+1}^{(p)} = U(C_t^{(p)}, R_t), \\
    I(C_{t+1}^{(p)}; X_t \mid R_t) \approx 0.
\end{aligned}    
\end{equation}
$U(\cdot)$ is the planner-state update function, and $I(A;B\mid C)$ denotes conditional mutual information. Eq.~\ref{eq:info-constraint} is not optimized as a differentiable objective. It is enforced by the interface design: raw execution traces are kept inside lower-level contexts and only schema-constrained returns are appended to the planner.

\subsection{HyMem Framework}
\label{sec:hymem-framework}

HyMem contains two context spaces: Planner, Executor, and two modules: isolated reasoning module, and Memory Management module. Only the planner persists across the whole task. Another spaces are instantiated on demand, return a typed message, and discard their private context.

\paragraph{Planner.}
The planner maintains the main decision context:
\begin{equation}
    C_t^{(p)} =
    \big(P^{(p)}, q, \mathcal{M}_t^{\mathrm{epi}},
    \mathcal{M}_t^{\mathrm{wm}}, \mathcal{M}_t^{\mathrm{tool}},
    H_t^{(p)}\big),
\end{equation}
where $\mathcal{M}^{\mathrm{epi}}$ is episode memory, $\mathcal{M}^{\mathrm{wm}}$ is working memory, $\mathcal{M}^{\mathrm{tool}}$ is compact tool memory, and $H_t^{(p)}$ stores only typed returns from previous sub-sessions. The planner emits exactly one action from:
\begin{equation}
\label{eq:action-set}
\begin{aligned}
    \mathcal{A} = \{&
    \langle\textsc{task}\rangle(d),
    \langle\textsc{isoReason}\rangle(d), \\
    &\langle\textsc{fold}\rangle,
    \langle\textsc{answer}\rangle(\hat{y})\}.
\end{aligned}
\end{equation}

\paragraph{Executor.}
Given a tool-use directive $d$, HyMem creates a fresh executor context:
\begin{equation}
    C_{t,0}^{(e)} = (P^{(e)}, d, \mathcal{M}_t^{\mathrm{tool}}).
\end{equation}
The executor performs a bounded tool-call loop and obtains raw observations $o_{t,1}, \dots, o_{t,k}$. These observations are not returned directly. HyMem first applies a relevance-conditioned distillation operator:
\begin{equation}
    \tilde{o}_{t,j} = \phi(o_{t,j} \mid d),
\end{equation}
where $\phi$ removes navigation text, duplicated passages, formatting artifacts, irrelevant snippets, and tool-error boilerplate while preserving named entities, dates, numerical values, source identifiers, evidence snippets, and answer candidates. The distilled observations are then synthesized into:
\begin{equation}
\begin{aligned}
R_t^{(e)} ={}& \langle\textsc{results}\rangle \\
& [\text{findings},\ \text{evidence},\ \text{sources}, \\
& \quad \text{status},\ \text{gaps}] \\
& \langle/\textsc{results}\rangle .
\end{aligned}
\end{equation}
Only $R_t^{(e)}$ is appended to the planner history. The executor context and raw observations are discarded after return.

\paragraph{Isolated reasoning.}
For sub-tasks requiring multi-hop analysis, hypothesis checking, or reconciliation of conflicting evidence, the planner emits $\langle\textsc{isoReason}\rangle(d)$. HyMem then creates:
\begin{equation}
\begin{aligned}
    C_{t,0}^{(\mathrm{iso})}
    = \big(&P^{(\mathrm{iso})},\ d,\ 
    \mathcal{M}_t^{\mathrm{epi}},\
    \mathcal{M}_t^{\mathrm{wm}},\\
    &\mathcal{M}_t^{\mathrm{tool}},\
    \tau(H_t^{(p)})\big).
\end{aligned}
\end{equation}
where $\tau(\cdot)$ selects the recent planner state needed to ground the sub-task. The isolated session may call the executor for additional evidence, but its intermediate reasoning trace remains private. It returns:
\begin{equation}
\begin{aligned}
R_t^{(\mathrm{iso})} ={}& \langle\textsc{return}\rangle \\
& [\text{conclusion},\ \text{supporting facts},\ \text{sources}, \\
& \quad \text{confidence } \kappa,\ \text{assumptions}] \\
& \langle/\textsc{return}\rangle .
\end{aligned}
\end{equation}
where $\kappa \in [0,1]$ is the self-estimated confidence. After return, the isolated context is discarded.

\paragraph{Structured memories.}
HyMem maintains three rewritable memories:
\begin{equation}
    \mathcal{M}_t =
    \{\mathcal{M}_t^{\mathrm{epi}},
      \mathcal{M}_t^{\mathrm{wm}},
      \mathcal{M}_t^{\mathrm{tool}}\}.
\end{equation}
Episode memory records completed milestones, verified facts, and important decisions. Working memory records the current objective, unresolved gaps, and next-step plan. Tool memory records effective tool-use patterns, failed queries, and tool-specific caveats. The fold operator rewrites all memories jointly:
\begin{equation}
    \mathcal{M}_{t+1}
    =
    \Phi(H_t^{(p)}, H_{\leq t}^{(e)}, H_{\leq t}^{(\mathrm{iso})}, \mathcal{M}_t).
\end{equation}

HyMem uses asymmetric memory injection:
\begin{equation}
\begin{aligned}
    C^{(p)} &\leftarrow
    (\mathcal{M}^{\mathrm{epi}}, \mathcal{M}^{\mathrm{wm}}, \mathcal{M}^{\mathrm{tool}}), \\
    C^{(\mathrm{iso})} &\leftarrow
    (\mathcal{M}^{\mathrm{epi}}, \mathcal{M}^{\mathrm{wm}}, \mathcal{M}^{\mathrm{tool}}), \\
    C^{(e)} &\leftarrow
    \mathcal{M}^{\mathrm{tool}}.
\end{aligned}
\end{equation}
The tool memory injected into the planner is a compact procedural summary, not raw tool logs. This preserves reusable tool experience while keeping raw execution traces outside the planner.

\subsection{Inference Procedure and Context Boundary}
\label{sec:hymem-procedure}

HyMem wraps a frozen base LLM $\pi_\theta$ at inference time. At planner step $t$:
\begin{equation}
    a_t \sim \pi_\theta(\cdot \mid C_t^{(p)}),
    \ a_t \in \mathcal{A}.
\end{equation}
The transition is:
\begin{equation}
\small
\begin{aligned}
a_t = \langle\textsc{task}\rangle(d)
&\Rightarrow
C_{t+1}^{(p)} \leftarrow
U\!\left(C_t^{(p)}, R_t^{(e)}\right), \\
a_t = \langle\textsc{isoReason}\rangle(d)
&\Rightarrow
C_{t+1}^{(p)} \leftarrow
U\!\left(C_t^{(p)}, R_t^{(\mathrm{iso})}\right), \\
a_t = \langle\textsc{fold}\rangle
\ \mathrm{or}\ \mathrm{Trigger}(s_t)
&\Rightarrow
\mathcal{M}_{t+1} \leftarrow \Phi(\cdot), \\
a_t = \langle\textsc{answer}\rangle(\hat{y})
&\Rightarrow
\mathrm{return}\ \hat{y}.
\end{aligned}
\end{equation}

A fold is invoked when the planner explicitly emits $\langle\textsc{fold}\rangle$, or when one of the following deterministic conditions is met:
\begin{equation}
\label{eq:trigger}
\begin{aligned}
\mathrm{Trigger}(s_t)
= {}& [r_t^{\mathrm{ctx}} > 0.99]
    \lor [n_t^{\mathrm{exec}} \geq 5] \\
& \lor [n_t^{\mathrm{fail}} \geq 3]
    \lor [n_t^{\mathrm{turn}} \geq 8] \\
& \lor [n_t^{\mathrm{tool}} \geq 10].
\end{aligned}
\end{equation}
Here, $r_t^{\mathrm{ctx}}$ is planner-context token utilization, $n_t^{\mathrm{exec}}$ counts executor invocations since the last fold, $n_t^{\mathrm{fail}}$ counts consecutive executor failures, $n_t^{\mathrm{turn}}$ counts planner turns since the last fold, and $n_t^{\mathrm{tool}}$ counts tool calls since the last fold.

The context boundary follows from the transition rules. Raw observations enter only the executor and are deleted after $R_t^{(e)}$ is returned. Intermediate reasoning tokens enter only the isolated reasoning context and are deleted after $R_t^{(\mathrm{iso})}$ is returned. The planner grows through bounded typed returns and rewritten memories, rather than through the full length of tool traces or sub-task deliberations. Therefore, HyMem does not merely summarize a flat trajectory after it becomes long; it restricts which information can enter the decision context throughout execution. This mechanism underlies the lower context-dilution behavior measured in Figure~\ref{fig:task_info}.

\section{Experiments}
\label{sec:experiments}


\begin{table*}[t]
    \centering
    \footnotesize 
    \renewcommand{\arraystretch}{1.3} 
    \setlength{\tabcolsep}{3.8pt}      
    
    \begin{tabular}{cc|cccc|cccc}
    \toprule
    \textbf{Backbone} & \textbf{Framework} & \multicolumn{4}{c|}{\textbf{GAIA}} & \multicolumn{4}{c}{\textbf{Browsecomp-plus}} \\
    & & Level 1 & Level 2 & Level 3 & \textbf{Avg. P@1} & Easy & Medium & Hard & \textbf{Avg. P@1}  \\
    \hline
    
    \rowcolor[HTML]{E2F0D9} \multicolumn{10}{c}{\textit{\textbf{No Agency}}} \\
    Qwen3-32B & Base & 23.8 & 6.0 & 5.3 & 18.9 & - & - & - & -  \\ \hline
    DeepSeek-V4 & Base & 35.7 & 27.2 & 15.8 & 28.3 & 14.0 & 10.0 & 0.0 & 8.0  \\
    
    \hline
    \rowcolor[HTML]{FCE4D6} \multicolumn{10}{c}{\textit{\textbf{Agentic Frameworks}}} \\
    \multirow{2}{*}{Qwen3-32B} & ReAct & 19.0 & 10.6 & 0.0 & 11.8 & 70.0 & 24.0 & 2.0 & 32.0  \\
     & Summary & 30.9 & 21.2 & 5.3 & 22.0 & 20.6 & 46.0 & 14.0 & 2.0  \\ \hline
    \multirow{2}{*}{DeepSeek-v4} & ReAct & 66.6 & 56.1 & 36.8 & 56.6 & \textbf{94.0} & 62.0 & 14.0 & 56.6 \\
     & Summary & 69.1 & 45.5 & 31.6 & 51.2 & 78.0 & 42.0 & 6.0 & 42.0 \\
     & ReSum & 76.2 & 54.5 & 47.4 & 60.6 & 78.0 & 46.0 & 6.0 & 43.3 \\
     & AMEM & 61.9 & 48.5 & 26.3 & 49.6 & 70.0 & 34.0 & 6.0 & 36.7 \\
     & THREAD & 42.9 & 34.8 & 26.3 & 35.4 & 42.0 & 30.0 & 8.0 & 26.7 \\
    
    \hline
    \rowcolor[HTML]{BDD7EE} \multicolumn{10}{c}{\textit{\textbf{Our Framework}}} \\
    Qwen3-32B & HyMem & 45.2 & 30.3 & 15.8 & 33.0 & 48.0 & 24.0 & 4.0 & 25.3  \\ \hline
    DeepSeek-V4 & HyMem & \textbf{77.0} & \textbf{65.2} & \textbf{49.1} & \textbf{66.7} & 90.0 & \textbf{64.0} & \textbf{30.0} & \textbf{61.3} \\
    \toprule
    \end{tabular}
    \caption{Experimental results on GAIA and Browsecomp-plus benchmarks. P@1 denotes Pass@1. The bolded numbers represent the top performers in their respective categories. (1) HyMem improves the reasoning capabilities of different backbone models. (2) HyMem delivers the largest overall performance improvement among the compared agent frameworks.}
    \label{tab:results}
\end{table*}

\noindent \textbf{Benchmark.} We evaluate HyMem using two benchmark datasets including GAIA~\citep{mialon2024gaia}, and Browsecomp-plus~\citep{chen2025BrowseCompPlus}. Browsecomp-plus focuses on reasoning-intensive queries derived from OpenAI's Browsecomp~\citep{wei2025browsecompsimplechallengingbenchmark} and provides a comprehensive offline corpus to ensure evaluation consistency without real-time web dependency. GAIA is employed to assess the agent’s proficiency in general reasoning, retrieval, and tool utilization. Due to constraints in retrieval API resources and computational time, we randomly sampled 150 questions with a balanced difficulty distribution from Browsecomp-plus~\citep{sun2025scalinglonghorizonllmagent}. For GAIA, the evaluation is restricted to text-based queries.

\noindent \textbf{Baseline and Implementation.} 
HyMem is compared against several representative methodologies: the \textbf{ReAct} framework~\citep{yao2023react}, which facilitates explicit reasoning via a continuous reason-act-observe cycle; \textbf{Summarization-based} workflows~\citep{yu2025memagentreshapinglongcontextllm}, which condense historical interaction trajectories into information-dense summaries to guide subsequent reasoning; \textbf{ReSum}~\citep{wu2025resumunlockinglonghorizonsearch}, which periodically summarizes long interaction histories into compact reasoning states to support continued long-horizon exploration; \textbf{A-MEM}~\citep{xu2025amemagenticmemoryllm}, which constructs an agentic memory system by organizing experiences into dynamically linked and evolving memory notes; and \textbf{THREAD}~\citep{schroeder-etal-2025-thread}, which decomposes complex tasks through recursive thread spawning, enabling isolated subtask reasoning and subsequent aggregation. We evaluate two backbone models, Qwen3-32B~\citep{yang2025qwen3technicalreport} and DeepSeek-V4~\citep{deepseekai2025deepseekv32pushingfrontieropen}. Unless otherwise specified, the same backbone is used for the compared framework under the same benchmark setting. Detailed prompts, turn budgets, and hyperparameters are provided in the Appendix~\ref{appendix：experiment_setup}.

\begin{figure*}[t]
\centering
\begin{subfigure}{0.48\textwidth}
    \centering
    \includegraphics[width=\linewidth]{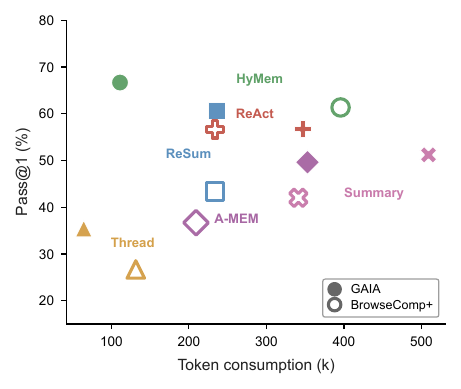}
    \caption{Token Consumption vs. Performance}
    \label{fig:token_vs_performance}
\end{subfigure}
\hfill 
\begin{subfigure}{0.48\textwidth}
    \centering
    \includegraphics[width=\linewidth]{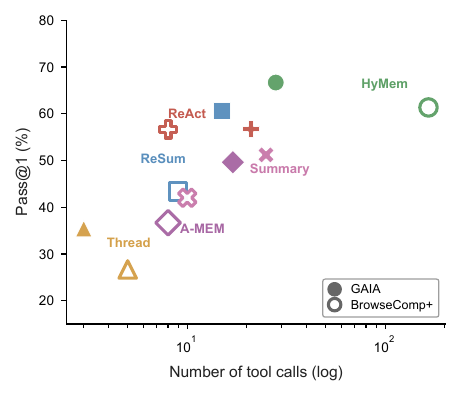}
    \caption{Tool Calls vs. Performance}
    \label{fig:tool_vs_performance}
\end{subfigure}
    \caption{\textbf{Resource Consumption and Task Performance across Different Methods.} We compare the resource consumption and task performance of ReAct, ReSum, A-MEM, THREAD, and HyMem on the GAIA and BrowseComp-Plus tasks. HyMem achieves better task performance while maintaining acceptable resource utilization.}
    \label{fig:overall_analysis}
\end{figure*}

\subsection{Results and Analysis}
Table~\ref{tab:results} shows that HyMem achieves the strongest overall performance when paired with the stronger DeepSeek-V4 backbone. On GAIA, DeepSeek-V4 with HyMem reaches an average Pass@1 of 66.7, improving over ReAct by 10.1 percentage points and over ReSum by 6.1 percentage points. The gain is consistent across all three GAIA difficulty levels, with the largest margin over ReAct appearing on Level 3 questions. On Browsecomp-plus, HyMem reaches an average Pass@1 of 61.3, outperforming ReAct by 4.7 percentage points and substantially exceeding memory-oriented or modular baselines such as ReSum, A-MEM, and THREAD.

The difficulty-wise results further clarify where HyMem is most useful. On Browsecomp-plus with DeepSeek-V4, HyMem is slightly lower than ReAct on easy questions, but improves on medium and hard questions, especially hard questions where Pass@1 increases from 14.0 to 30.0. This pattern is consistent with the design motivation of HyMem: typed context isolation is most beneficial when solving the task requires sustained planning, evidence accumulation, and multi-step verification rather than short tool-use trajectories.

The Qwen3-32B results show a more nuanced picture. On GAIA, HyMem improves the average Pass@1 from 11.8 under ReAct and 22.0 under the summarization workflow to 33.0, indicating that context isolation can help a smaller backbone sustain reasoning over multi-step tasks. However, on Browsecomp-plus, Qwen3-32B with HyMem does not outperform the ReAct baseline. This suggests that HyMem's gains depend not only on context structure but also on the underlying model's ability to follow typed control instructions and produce reliable structured returns. We therefore treat cross-backbone robustness as promising but not uniform across all benchmark settings.

\begin{figure}[t]
    \includegraphics[width=\columnwidth]{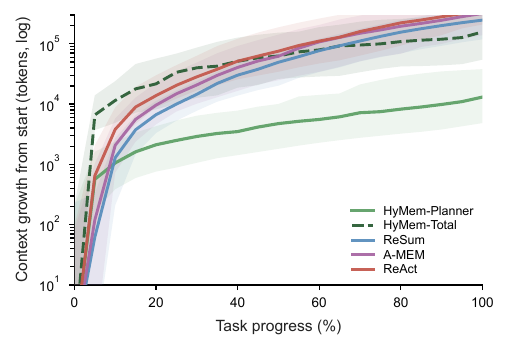}
    \caption{\textbf{Relationship between Context Changes and Task Progress across Different Methods.} We record the relationship between context growth trends and task progress for ReAct, A-MEM, ReSum, and HyMem. HyMem-Total denotes the overall context of HyMem, while HyMem refers to the context of the planner module.}
    \label{fig:context_token}
\end{figure}

\subsection{Efficiency and Context Behavior}

Figure~\ref{fig:overall_analysis} compares task performance against token consumption and tool-call count. HyMem occupies a favorable region of the performance-resource trade-off: it achieves higher task performance than ReAct, ReSum, A-MEM, and THREAD while keeping resource usage within a comparable range. This is important because HyMem introduces additional control structure, including isolated reasoning and memory folding. The analysis indicates that the performance gain is not simply obtained by unbounded context growth or excessive tool use; rather, structured context isolation directs additional computation toward task-relevant reasoning and evidence synthesis.

Figure~\ref{fig:context_token} shows that HyMem’s advantage lies in how context is allocated rather than in reducing total context usage. HyMem-Total continues to grow as the Executor and Isolated Reasoning modules perform tool interaction and multi-step analysis, whereas the persistent Planner context grows much more slowly. The separation between the two curves indicates that most execution details and intermediate analysis remain within temporary contexts, while only structured returns and rewritten memories update the Planner. Compared with methods whose persistent reasoning contexts grow with accumulated interaction traces, HyMem supports extensive exploration while keeping the Planner context focused on task-relevant information and preventing transient execution details from accumulating throughout the task.

\begin{table*}[t]
    \centering
    \footnotesize
    \renewcommand{\arraystretch}{1.4} 
    \setlength{\tabcolsep}{5pt}       

    \begin{tabular}{l|ccc|ccc}
    \hline
    \multirow{2}{*}{\textbf{Framework}} & \multicolumn{3}{c|}{\textbf{Browsecomp-plus (DeepSeek-V4)}} & \multicolumn{3}{c}{\textbf{GAIA (Qwen3-32B)}} \\
    & Pass@1  & Plan. & Conf. & Pass@1  & Plan. & Conf. \\ \hline

    \textbf{HyMem (Full)} & \textbf{0.6300}  & 12.98 & 0.77 & \textbf{0.3200}  & 14.20 & 0.82 \\ \hline
    
    w/o Isolated Reasoning & 0.5800
    \negat{5.00pp}  & 10.26\negat{2.72} & 0.68\negat{0.09} & 0.1340 \negat{18.60pp}  & 11.50\negat{2.70} & 0.71\negat{0.11} \\ 
    w/o Memory Management & 0.5400 \negat{9.00pp}  & 9.68\negat{3.30} & 0.69\negat{0.08} & 0.1340 \negat{18.60pp}  & 10.12\negat{4.08} & 0.70\negat{0.12} \\ \hline

    \end{tabular}
    \caption{Ablation study of HyMem. Red values \negat{} indicate the performance drop compared to the full framework.}
    \label{tab:ablation}
\end{table*}

\subsection{Ablation Study}

Table~\ref{tab:ablation} evaluates the contribution of the two main components of HyMem: the Isolated Reasoning Module and Memory Management. Removing isolated reasoning causes the large performance drop, reducing Pass@1 by 5.0 percentage points on Browsecomp-plus and 18.6 percentage points on GAIA. This confirms that isolating deep sub-task reasoning is crucial for long-horizon agent tasks, especially when the main planner would otherwise accumulate long intermediate deliberations.

Removing memory management also substantially degrades performance, reducing Pass@1 by 9.0 percentage points on Browsecomp-plus and 18.6 percentage points on GAIA. This result shows that typed isolation alone is insufficient: the agent also needs structured memories to preserve milestones, current goals, and reusable tool experience across context refreshes. Together, the ablation results support the two-part design of HyMem. Isolated reasoning protects the planner from sub-task deliberation noise, while memory management preserves task continuity after histories are folded.

Finally, the auxiliary metrics in Table~\ref{tab:ablation} help interpret the failure modes. Without isolated reasoning, Pass@1 decreases on both benchmarks, indicating that the agent is less effective at resolving complex subtasks when their analysis is handled directly within the persistent planning context. Without memory management, confidence also drops, indicating that losing structured task state makes the agent less certain even when it can still execute tools. These trends align with the proposed mechanism: HyMem improves long-horizon performance by coordinating where reasoning happens and how task state is retained.

\section{Conclusion}
\label{sec:conclusion}

This paper addresses context dilution in long-horizon LLM agents, where sparse planning-relevant signals are gradually obscured by dense execution traces. We proposed HyMem, a training-free inference-time framework based on typed context isolation. By separating planning, tool execution, isolated sub-task reasoning, and memory consolidation into distinct context spaces, HyMem restricts raw observations and intermediate deliberation traces from directly entering the main planning context. Structured returns and asymmetric memory injection allow the agent to preserve task continuity while keeping the planner focused on verified facts, current goals, and unresolved gaps.

Experiments on GAIA and Browsecomp-plus show that HyMem improves long-horizon agent performance most consistently with a stronger instruction-following backbone. With DeepSeek-V4, HyMem achieves the best average Pass@1 on both benchmarks and shows clear gains on harder Browsecomp-plus questions. The resource analysis further suggests that these gains are not simply obtained by using more tokens or more tool calls, while the context-growth analysis supports the intended mechanism: HyMem allows execution and isolated reasoning contexts to expand without proportionally expanding the persistent planner context. Ablation results confirm that both isolated reasoning and structured memory management are necessary for the full framework.

HyMem also has clear boundaries. Its effectiveness depends on the model's ability to follow typed control instructions and produce reliable structured returns, and the gains are not uniform across all backbone and benchmark combinations. Future work should study adaptive routing and memory-folding policies, reduce the cost of auxiliary distillation, and extend context isolation to interactive settings where user intent may change during multi-turn task execution.

\section*{Limitations}
LLM-based agents are increasingly used for long-horizon tasks that require iterative planning, tool use, and information gathering. However, their performance is often limited by context dilution: as interaction histories grow, dense execution traces such as raw tool outputs, retries, and intermediate deliberations can obscure the sparse planning signals needed for global task control. Existing compression and retrieval methods reduce context length, but they usually operate after planning and execution information has already been mixed into a flat trajectory. We propose HyMem, a training-free inference-time framework based on typed context isolation. HyMem separates planning, tool execution, isolated sub-task reasoning, and memory consolidation into distinct context spaces, allowing only schema-constrained results and structured memories to enter the main planner context. Experiments on GAIA and Browsecomp-plus show that HyMem improves performance over ReAct, ReSum, and other agent baselines while maintaining an acceptable resource-performance trade-off.

\bibliography{reference}

\appendix

\begin{table}[h]
    \centering
    \small
    \renewcommand{\arraystretch}{1.3}
    \begin{tcolorbox}[colback=white, colframe=black!70, boxrule=0.8pt, arc=3pt, left=6pt, right=6pt, top=6pt, bottom=6pt]
    
        \begin{tcolorbox}[colback=black!5, colframe=white, left=2pt, right=2pt, top=2pt, bottom=2pt]
        \textbf{User Query:} Born in 1970s, 3x World Half-Marathon champ... retired 2021. \textbf{Goal:} Date (DD/MM/YYYY) of 10,000m PB?
        \end{tcolorbox}
        \vspace{0.5em}
    
        \textbf{Turn 1: Entity Identification} \\
        \think{Break down constraints: "Born 1970s", "3x Champion". First, identify the athlete.(\dots)} \\
        \task{Search "three-time World Half-Marathon champion born 1970s"}
        \result{Target matches \textbf{Tegla Loroupe} (Kenyan). Born May 9, 1973. Won titles 1997, 1998, 1999. (...)}
    
        \vspace{0.4em}
        \hrule
        \vspace{0.4em}
    
        \textbf{Turn 2-4: Constraint Verification} \\
        \think{Candidate found. Before searching for PB, verify detailed clues (weight, UN honor, retirement) in parallel.} \\
        \branch{Verify: (1) First win age 21 (NYC Marathon '94?); (2) UN Person of Year; (3) Weight 40kg.}
        \result{\textbf{[Success]} Confirmed: Won NYC Marathon at 21; UN Person of Year (Kenya, 2016); Weight approx 40kg.}
    
        \vspace{0.4em}
        \hrule
        \vspace{0.4em}
    
        \textbf{Turn 5: Attribute Retrieval} \\
        \think{Identity verified. Now search specifically for the 10,000m personal best date.} \\
        \task{Search "Tegla Loroupe 10,000 meters personal best date"}
        \result{Found PB time: \textbf{30:32.03}. \textbf{[Failure]} Specific date missing in snippets. Source only lists time.}
    
        \vspace{0.4em}
        \hrule
        \vspace{0.4em}
    
        \textbf{Turn 6: Refinement Strategy} \\
        \think{Direct search for date failed. Strategy shift: Search using the exact time "30:32.03" to find the specific event record.} \\
        \task{Search "Tegla Loroupe 30:32.03 10,000m"}
        \result{Event Identified: \textbf{1999 World Championships (Sevilla)}. Date confirmed: \textbf{26 August 1999}.}
    
        \memcompress{Detailed traces (Turns 1-6) folded into \textbf{Assignment, Moving \& Tool Memory}. \\ Updated Memory: \textit{"Tegla Loroupe, PB 30:32.03 on 26 Aug 1999 (Needs Final Verify)"}.}
    
        \vspace{0.2em}
        \textbf{Turn 7-8: Final Answer} \\
        \think{Memory confirms date. Evidence is consistent across sources (Sporting-heroes, Wikipedia).} \\
        \textbf{Answer:} Tegla Loroupe achieved her personal best for the 10,000-meter race on \textbf{26 August 1999}.
    
    \end{tcolorbox}
    \caption{Execution trace of HyMem on a multi-hop reasoning task. The flow demonstrates the interaction between the \textcolor{hy_planner}{Planner Layer} (reasoning/task generation), \textcolor{hy_exec}{Executor Layer} (tool use), and the \textcolor{hy_branch}{Isolated Reasoning} mechanism for verification. Note the \textcolor{hy_mem}{Memory Compression} event which consolidates the context before the final answer.}
    \label{table:case}
\end{table}

\section{Experiment setup}
\label{appendix：experiment_setup}

\subsection{Benchmarks and Data Selection}
For the primary evaluation, we utilized a curated subset of 150 tasks from the BrowseComp-Plus benchmark, maintaining a balanced 1:1:1 ratio across easy, medium, and hard difficulty levels. For the GAIA benchmark, we focused on the validation set (165 tasks), specifically filtering for and evaluating all text-only instances. In ablation studies, we maintained the same 150 tasks for BrowseComp-Plus and the text-only GAIA subset for testing. All experimental results are reported as the mean value derived from at least three independent trials to ensure statistical reliability and consistency.

\subsection{Model Configurations \& Hyperparameters}
The agent frameworks were evaluated using Qwen3-32B and DeepSeek-V4 as the primary backbone models. To ensure procedural consistency and experimental stability, all auxiliary language modeling tasks across all frameworks were uniformly handled by DeepSeek-V4.

To regulate task progression and optimize computational efficiency, the HyMem framework employs a specific set of hyperparameter constraints:
\begin{itemize}[nosep]
    \item \textbf{\texttt{Planner\_turns=12}}: The maximum number of interaction turns permitted within the \textit{planner Layer} for high-level strategic planning.
    \item \textbf{\texttt{execute\_turns=10}}: The maximum allowable turns for the \textit{Actor Layer} during discrete task execution and tool manipulation.
    \item  \textbf{\texttt{max\_iso\_reasoning=5}}: The upper limit on the total number of \textit{isolated reasoning sessions} invoked throughout the duration of a task.
    \item \textbf{\texttt{max\_iso\_reasoning\_turn=3}}: The maximum number of reasoning steps allowed within each individual \textit{isolated reasoning session}.
\end{itemize}
This configuration was selected to ensure the agent has sufficient latitude for environmental exploration and complex decision-making while preventing excessive task duration or inefficient resource consumption. We compared HyMem against two standard baselines. 
ReAct maximum turn budget was set to 100. Summarization-based Agent: When the context length reached a predefined threshold, the agent performed recursive summarization by distilling the most recent three interaction turns and concatenating them with the preceding history. This baseline was also capped at 100 turns.

\subsection{Context Analysis}
To evaluate global context management efficiency, we collected historical trajectories from identical tasks for both HyMem, ReAct, ReSum and A-MEM. Figure~\ref{fig:context_token} plots task progress on the horizontal axis and context volume on the vertical axis. To visualize the aggregate effect, we concatenated the context-change data from all evaluated tasks into a continuous curve, providing a holistic comparison of how each framework manages information density over time.

\section{Case Study}

To demonstrate the effectiveness of the HyMem framework in leveraging hierarchical isolation for context management to enhance reasoning capabilities in complex, long-horizon tasks, Table~\ref{table:case} provides a detailed case study. This instance illustrates how the Planner and Executor layers utilize architectural decoupling to filter redundant information while facilitating the exchange of high-value strategic signals. Furthermore, it highlights how the isolated reasoning module and memory management mechanism synergistically enhance exploratory steps and distill essential information to maintain reasoning integrity, ultimately enabling the resolution of intricate multi-hop tasks. For the sake of clarity, the content in the table has been selectively curated and condensed to emphasize the specific roles of these core components and provide a transparent illustration of the framework’s operational mechanics.

\section{Prompt Engineering}
\label{appendix:prompts}

The primary prompt templates for the HyMem framework are detailed in Tables~\ref{table:coord_prompt} --~\ref{table:memo_prompt}, representing the Planner Layer, Executor Layer, isolated reasoning sessions, and memory management components, respectively~\citep{sun2025scalinglonghorizonllmagent,li2025memosoperatingmemoryaugmentedgeneration}.

\begin{table*}[t]
    \centering
\begin{exmp}{HyMem Planner Layer}{}
 \scriptsize
You are an expert research planner with memory-augmented reasoning capabilities. Your task is to analyze complex questions and devise effective search strategies. You work with a specialized search \textbf{actor} and can create reasoning isoReasons. Your goal is to reach the best final answer by combining structured analysis, targeted retrieval, delegated reasoning, and memory-guided iteration. \\
\textbf{\#\# Role \& Responsibilities} \\
Your role is to: \\
- Analyze the problem and identify key information needs. \\
- Decompose complex questions into concrete search tasks and reasoning sub-problems. \\
- Assign search tasks to the \textbf{actor} or delegate complex reasoning to isoReasons. \\
- Synthesize results and reason towards the final answer. \\
\textbf{\#\# Workflow} \\
Follow this process iteratively: \\
\textbf{(a) Problem Analysis}: Identify core entities, constraints, and relationships. \\
\textbf{(b) Search Strategy}: Design strategies that narrow the search space and reduce ambiguity.\\ 
\textbf{(c) Task Delegation}: You have \textbf{two} delegation options: \\
- \textbf{Option A -- Search Task (for actor)}: 
\begin{verbatim}
<task> [Describe the specific search task for the actor.] </task>
\end{verbatim}
- \textbf{Option B -- isoReason Task (independent reasoning)}: 
\begin{verbatim}
<isoReason>
  <description> [3-5 word identifier] </description>
  <prompt> [Clear reasoning objective] </prompt>
</isoReason>
\end{verbatim}
\textbf{(d) Result Integration}: When the \textbf{actor} or \textbf{isoReason} returns results: \\
- Evaluate whether results satisfy the criteria. \\
- Identify remaining gaps and decide whether to continue exploration or conclude. \\
\textbf{(e) Memory-Guided Reasoning}: Use memory from previous sessions to: \\
- Avoid repeating failed strategies. \\
- Build upon prior discoveries. \\
- Maintain reasoning continuity across iterations. \\
\textbf{\#\# Output Format} \\
Your response must include step-by-step reasoning and a final answer: 
\begin{verbatim}
<think>
Step-by-step reasoning, including problem analysis, strategy design,
delegation choices, and integration decisions.
If reasoning becomes too lengthy, output <to_memory> to trigger memory compression.
</think>
<answer> [Your final answer] </answer>
<confidence> [Score 0-1] </confidence>
\end{verbatim}
\textbf{\#\# Important Guidelines} \par
1. Think step-by-step within \texttt{<think>...</think>} tags. \\
2. Use \textbf{isoReason} when a sub-problem needs independent analysis. \\
3. Use \texttt{<task>} for straightforward information retrieval. \\
4. If reasoning becomes too lengthy, output \texttt{<to\_memory>} to trigger memory compression. \\
5. Do not stop reasoning until you can provide a final answer. \\
6. If stuck, reanalyze the problem from different angles and revise the search plan. \\

\end{exmp}
    \caption{Full prompt template for HyMem Planner Layer.}
    \label{table:coord_prompt}
\end{table*}

\begin{table*}[t]
    \centering
\begin{exmp}{HyMem Executor Layer}{}
\scriptsize
You are a searching agent equipped with multiple search tools and file parsing capabilities. Your task is to locate, extract, and synthesize information that satisfies a given search target, using careful query design and iterative reasoning. \\
\\
\textbf{\#\# Role \& Objective} \\
Your goal is to efficiently find accurate and relevant information by: \\
- Designing precise search queries. \\
- Selecting appropriate tools for retrieval or parsing. \\
- Extracting structured evidence from web pages or files. \\
- Synthesizing findings into clear results aligned with the search target. \\
\\
\textbf{\#\# Available Tools} \\
You may use the following tools at any step: 
\begin{itemize}[nosep]
    \item \texttt{web\_search(keywords)}: Perform keyword-based web searches. 
    \item \texttt{web\_parse(link, query)}: Fetch a specific web page and extract information relevant to a query. 
    \item \texttt{batch\_search\_and\_filter(keyword)}: Run integrated searches with built-in filtering. 
    \item \texttt{parse\_file(file\_path)}: Parse and extract text from files (PDF, CSV, XLSX, DOCX, TXT, HTML).
\end{itemize}
\textbf{\#\# Workflow} \\
Follow this structured process: \\
\textbf{(a) Target Analysis}: Clearly understand the search target and define information requirements. \\
\textbf{(b) Query Design}: Craft focused keywords; avoid overly broad searches. \\
\textbf{(c) Tool Selection}: Choose the most suitable tool for each subtask (web search, page parsing, or file parsing). \\
\textbf{(d) Iterative Exploration}: Refine queries or parse deeper sources based on intermediate findings. \\
\textbf{(e) Result Synthesis}: Aggregate extracted evidence into concise, relevant findings. \\
\\
\textbf{\#\# Example Usage} 
\begin{verbatim}
<code>
# Web search
result = await web_search("machine learning papers 2023")
for item in result["tool_result"]["organic"][:5]:
    print(f"- {item['title']}: {item['snippet'][:100]}")

# File parsing
file_result = await parse_file("/path/to/document.pdf")
content = file_result["tool_result"]["content"]
print(f"File content: {content[:500]}")
</code>
\end{verbatim}

\textbf{\#\# Output Format}
Your response must include reasoning and final findings:
\begin{verbatim}
<think>
Step-by-step reasoning about query selection, tool usage,
evidence evaluation, and refinement decisions.
</think>
<results>
[Structured list or summary of findings that satisfy the search target.]
</results>
\end{verbatim}

\textbf{\#\# Guidelines}

1. Always reason explicitly within \texttt{<think>...</think>} tags. \\
2. Avoid unnecessary or overly broad searches. \\
3. Use \texttt{web\_parse} for detailed extraction from promising sources. \\
4. Use \texttt{parse\_file} when handling provided documents. \\
5. Iterate until the search target is sufficiently satisfied, then report results clearly. \\

\end{exmp}
    \caption{Full prompt template for HyMem Executor Layer. }
    \label{tabled:act_prompt}
\end{table*}

\begin{table*}[t]
    \centering
\begin{exmp}{HyMem Isolated Reasoning Module}{}
\scriptsize
You are a reasoning branch of the main planner. You inherit the full context and memory from \textbf{MAIN}. Your role is to focus exclusively on the assigned reasoning task, execute systematic analysis, optionally gather information, and return immediately after completing the objective. You must not perform any actions beyond the specified scope. \\
\\
\textbf{\#\# Role \& Scope} \\
Your responsibilities are strictly limited to: \\
- Concentrating on the assigned reasoning objective only. \\
- Applying structured, step-by-step analysis. \\
- Returning comprehensive findings to \textbf{MAIN} without attempting to solve the overall problem. \\
\\
\textbf{\#\# Critical Instructions} \\
You must follow these rules precisely: 
\begin{enumerate}[nosep]
    \item \textbf{Understand the Task}: Carefully read and internalize the assigned objective. 
    \item \textbf{Execute Analysis}: 
    \begin{itemize}[nosep]
        \item Perform step-by-step reasoning within \texttt{<think>...</think>} tags. 
        \item Use \texttt{<task>...</task>} to delegate searches to the executor if external information is required. 
        \item Analyze all evidence systematically and critically. 
    \end{itemize}
    \item \textbf{Return Findings}: Once the objective is complete, immediately report results using the required return format. 
\end{enumerate}

\textbf{\#\# Return Format Requirements} \\
Because your internal reasoning is invisible to \textbf{MAIN}, your return must be explicit, structured, and self-contained. Include all of the following components: 
\begin{itemize}[nosep]
    \item \textbf{CONCLUSION}: A direct answer to the assigned question, with concise justification. 
    \item \textbf{KEY EVIDENCE}: Supporting facts, observations, or data points (include sources or doc IDs if available). 
    \item \textbf{CONFIDENCE}: Your confidence in the conclusion (scale 0--1). 
    \item \textbf{NOTES}: Caveats, limitations, assumptions, or recommendations for \textbf{MAIN}. 
\end{itemize}
\textbf{\#\# Operational Constraint} \\
- Focus \textbf{exclusively} on the assigned task. \\
- Do \textbf{NOT} provide a final answer to the main problem. \\
- Do \textbf{NOT} perform actions outside your designated reasoning scope. \\
\\
\textbf{\#\# Output Format} \\
When you have completed your assigned reasoning task, wrap everything strictly in the following structure: 
\begin{verbatim}
<think>
Step-by-step reasoning specific to the assigned task.
</think>
<return>
CONCLUSION: ...
KEY EVIDENCE: ...
CONFIDENCE: ...
NOTES: ...
</return>
\end{verbatim}
\end{exmp}
    \caption{Full prompt template for the HyMem Isolated Reasoning Module, responsible for recursive sub-problem decomposition and independent analysis.}
    \label{table:isorea_prompt}
\end{table*}

\begin{table*}[t]
    \centering
\begin{exmp}{HyMem Memory Management Module}{}
\scriptsize
You are a memory management assistant responsible for compressing, structuring, and recording different forms of agent memory. Your objective is to distill complex interaction histories into concise, machine-readable memory representations that support long-term reasoning, short-term planning, and tool-use optimization. \\
\textbf{\#\# Memory Modules Overview} \\
You operate three distinct memory modules, each with a clearly defined role, input scope, and output schema. You must strictly follow the instructions for each module and output \textbf{only} the specified JSON format. \\
\textbf{\#\# Module 1: episode Memory} \\
\textbf{Role}: You are a memory compression assistant. Summarize the key events in the planner’s reasoning process. \\
\textbf{Task}: \texttt{\{question\}} \\
\textbf{Inputs}: \texttt{\{planner\_history\}}, \texttt{\{executor\_summary\}} \\
\textbf{Instructions}: 
\begin{enumerate}[nosep]
    \item Identify major milestones, strategic decisions, and key discoveries. 
    \item Extract only critical events that inform long-term progress. 
    \item Output \textbf{only} the JSON format specified below. 
\end{enumerate}
\textbf{Output Schema (JSON)}
\begin{verbatim}
{
  "task_description": "Summary of the task and overall goals",
  "key_events": [
    { "step": "N", "description": "What was done", "outcome": "learned" }
  ],
  "current_progress": "Summary of progress and remaining work"
}
\end{verbatim}

\tcbline

\textbf{\#\# Module 2: working Memory} \\
\textbf{Role}: You are a working memory manager. Create a snapshot of the current research state. \\
\textbf{Inputs}: \texttt{\{recent\_history\}}, \texttt{\{hypothesis\_section\}} \\
\textbf{Instructions}: 
\begin{enumerate}[nosep]
    \item Extract \textbf{only} immediate goals, current challenges, and planned next steps. 
    \item Focus on actionable information and ignore completed work. 
    \item Output \textbf{only} the JSON format specified below. 
\end{enumerate}
\textbf{Output Schema (JSON)}
\begin{verbatim}
{
  "immediate_goal": "What the planner is currently determining",
  "current_challenges": "Main obstacles or gaps in information",
  "active_hypotheses": ["hypothesis1", "hypothesis2"],
  "next_actions": [
    { "type": "search", "description": "Action" }
  ]
}
\end{verbatim}
\tcbline
\textbf{\#\# Module 3: Tool Memory} \\
\textbf{Role}: You are a tool experience recorder. Synthesize tool usage patterns across interactions. \\
\textbf{Input}: \texttt{\{history\_str\}} \\
\textbf{Instructions}: 
\begin{enumerate}[nosep]
    \item Analyze which tool calls succeeded versus failed. 
    \item Identify effective search strategies and parameter patterns. 
    \item Output \textbf{only} the JSON format specified below.
\end{enumerate}
\textbf{Output Schema (JSON)}
\begin{verbatim}
{
  "tools_used": [
    {
      "tool_name": "str",
      "call_count": N,
      "effective_queries": []
    }
  ],
  "derived_rules": ["Rule learned from experience"],
  "useful_sources": ["URLs or docids"]
}
\end{verbatim}
\textbf{\#\# Global Constraints}
\begin{itemize}[nosep]
    \item Do \textbf{not} include any text outside the required JSON schemas.
    \item Do \textbf{not} mix content across memory modules.
    \item Ensure outputs are concise, structured, and suitable for long-term agent memory storage. 
\end{itemize}

\end{exmp}
    \caption{The suite of prompts used in the Memory Management Module, including episode memory, working memory, and Tool memory.}
    \label{table:memo_prompt}
\end{table*}

\end{document}